\documentclass[acmlarge]{acmart}
\makeatletter
\newcommand{\myconfshort}{\acmConference@shortname}
\newcommand{\myconffull}{\acmConference@name}
\newcommand{\myconfdate}{\acmConference@date}
\newcommand{\myconfloc}{\acmConference@venue}
\AtBeginDocument{
  \fancypagestyle{firstpagestyle}{
    \fancyhead{}%
    \fancyfoot[C]{}%
  }
  \fancyhf{}
  \fancyhead[LO]{\@headfootfont\shorttitle}%
  \fancyhead[RE]{\@headfootfont\@shortauthors}%
  \fancyhead[LE]{\@headfootfont\footnotesize \myconfshort, \myconfdate, \myconfloc}%
  \fancyhead[RO]{\@headfootfont\footnotesize \myconfshort, \myconfdate, \myconfloc}%
  \fancyfoot[C]{}%
}
\makeatother
\acmBooktitle{\conffull\@ (\confshort), \confdate, \confloc}
\AtBeginDocument{%
  }
\usepackage{graphicx}
\usepackage{listings}
\copyrightyear{2026}
\acmYear{2026}
\setcopyright{cc}
\setcctype{by}
\acmConference[FAccT '26]{The 2026 ACM Conference on Fairness, Accountability, and Transparency}{June 25--28, 2026}{Montreal, QC, Canada}
\acmBooktitle{The 2026 ACM Conference on Fairness, Accountability, and Transparency (FAccT '26), June 25--28, 2026, Montreal, QC, Canada}
\acmDOI{10.1145/3805689.3812398}
\acmISBN{979-8-4007-2596-8/2026/06}

\usepackage{needspace}
\usepackage{placeins} %
\usepackage{caption} %
\usepackage[utf8]{inputenc}
\usepackage[whole]{bxcjkjatype}
\begin{document}

\title{What Do People Actually Want From AI? Mapping Preference Plurality}
\author{Julia Sepúlveda Coelho}
\affiliation{%
  \institution{Oxford Internet Institute, University of Oxford}
  \city{Oxford}
  \country{United Kingdom}}
\email{julia.sepulveda@oii.ox.ac.uk}

\author{Scott A. Hale}
\affiliation{%
  \institution{Oxford Internet Institute, University of Oxford}
  \city{Oxford}
  \country{United Kingdom}
  }

  \affiliation{%
  \institution{Meedan}
  \city{San Francisco}
  \country{United States}
  }
\email{scott.hale@oii.ox.ac.uk}

\begin{abstract}
Large Language Models (LLMs) are often fine-tuned through Reinforcement Learning from Human Feedback (RLHF) to align with people's preferences and values. However, this method has known limitations: it aggregates conflicting preferences, often relies on unrepresentative samples, and uses only binary comparisons. Analysing 1,500 open-ended responses from the PRISM dataset across 75 countries, we examine what people actually want from AI systems and reveal concrete failures of current methods.

We find that different people want different things: most values are requested by fewer than a quarter of respondents, with truthfulness the sole exception at 49\%. Furthermore, the same words hide divergent meanings: when people describe what they mean by ``truthfulness'', they reveal distinct, potentially incompatible, epistemological bases, as some ask for sourced claims, some for expert opinions, and some even ask for unpopular views. Certain capabilities, namely how human-like a model behaves, and some features, like AI guardrails, are outright controversial, with some desiring them and others rejecting them. We additionally find that people often use contextual distinctions (what AI should do ``by default'' versus ``if requested'') that binary comparisons cannot capture.

These findings expose fundamental problems in current alignment practices. When 49\% request truthfulness but define it differently, this is unlikely to be captured by a single reward model. The persistence of high hallucination rates in well-funded models, despite users' clear demands for accuracy, suggests that current methods fail to identify actual preferences. This paper sheds light on the situated, contested, imperfect signals that are currently being flattened into universal preference models, a practice others have characterised as epistemic violence.
\end{abstract}



\received{13 January 2026}

\maketitle

\section{Introduction}

Social media platforms and search engines presented themselves as neutral, universal intermediaries for people and information~\cite{facebook_facebooks_2015,twitter_twitter_2022,google_our_nodate}. Over time, however, it became clear that content moderation and algorithmic ranking were far from neutral or purely technical, embedding consequential value judgments about what content to amplify and what harms to prevent~\cite{gillespie_trust_2023,noble_algorithms_2018}. It was, however, too late: platforms had concentrated immense private power, becoming inescapable through network effects~\cite{lehdonvirta_cloud_2022}, and gaining potential to strengthen or destabilize democracies~\cite{lorenz-spreen_systematic_2022}. This ultimately placed them at the centre of multiple antitrust and digital safety legal battles worldwide~\cite{booth_meta_2025,european_commission_commission_2025,european_commission_commission_2025-1,li_regulatory_2024,office_of_public_affairs_department_2025}.

AI systems risk repeating this trajectory. The current generation of large language models is shaped through alignment---the process of training models to behave according to human preferences and values~\cite{ji_ai_2025}. Just as content moderation decisions determine what billions of users see on social media, alignment decisions determine how AI systems respond to queries, refuse requests, and make recommendations. And just as social media companies framed these choices as neutral, AI labs claim to align with underspecified ``human values''~\cite{openai_our_2024} or the ``helpful, honest, and harmless'' (HHH) framework~\cite{anthropic_alignment_nodate}. This leaves the companies to operationalize what are essentially ``empty signifiers''~\cite{kirk_empty_2023,varshney_decolonial_2024}.

In practice, this operationalisation typically relies on aggregating human preferences through Reinforcement Learning from Human Feedback (RLHF). Although RLHF is the dominant paradigm for alignment, it is increasingly criticized on several fronts. From a methodological perspective, the approach is hindered by unrepresentative sampling, systemic data flaws, and modelling inaccuracies~\cite{ji_ai_2025,conitzer_position_2024,kirk_past_2023}. Furthermore, its application has been found to introduce cultural biases and encourage undesirable features such as sycophancy and sandbagging~\cite{tao_cultural_2024,perez_discovering_2022}.

In this work, we extend the critique of current alignment paradigms by investigating human values and preferences through more fine-grained, open-ended data. Through a mixed-methods analysis of the diverse PRISM dataset, utilizing qualitative coding and regression analysis, we find tensions and nuances that are neglected by binary preference models. Specifically, we demonstrate that even apparently consensual values, like truthfulness, hide varying and conflicting definitions, and that other preferences, like how human-like a model behaves, or how strict AI guardrails should be, are outright controversial. Our findings provide empirical evidence that RLHF fails to capture the true complexity of user intent and allows minority preferences to be subsumed by the majority consensus.

\section{Related work}
\subsection{AI alignment}

LLM training can be broadly split into two stages: pre-training and post-training. Pre-training consists of training a model to predict tokens using large corpora of data, and the resulting base model is a document generator, reflecting the biases inherent in the data. Post-training is what transforms this document generator into a conversational assistant and aligns it to human preferences and values. This can be achieved through many methods, although most of them use human feedback, varying in their preference sources, elicitation formats, and modelling approaches~\cite{jiang_survey_2025}. 

Most model providers heavily depend on Reinforcement Learning from Human Feedback (RLHF) for alignment~\cite{ji_ai_2025,conitzer_position_2024,lindstrom_ai_2024} although other techniques like DPO are also used and there is little transparency about the process. RLHF consists of collecting human preferences through pairwise comparisons, training a preference model on those responses, and using this preference model as a signal for reinforcement learning. It has been successful at making models generate responses that users prefer and reducing harmful outputs~\cite{ouyang_training_2022}. Another approach used by a major model provider is Constitutional AI~\cite{bai_constitutional_2022}, which is classified as Reinforcement Learning from AI Feedback (RLAIF).

RLHF, however, has been widely criticised on both technical and theoretical points~\cite{kirk_past_2023,ji_ai_2025,lindstrom_ai_2024,casper_open_2023,lambert_alignment_2024}, and some of the criticism can be extended to Constitutional AI. The first problem is simply: whose human feedback?

\subsection{Whose values, whose preferences?}

The data necessary for RLHF has often been created by unrepresentative samples of the global population, often drawn from WEIRD demographics including crowdworkers, tech workers and university students~\cite{kirk_past_2023}. However, different people and different communities have different values, and as such may want to shape AI differently~\cite{sutrop_challenges_2020,han_cross-cultural_2025}. 

This problem also applies to some degree to Constitutional AI, which relies on a short list of principles. Generally, this list of principles is not participative. It draws on the UN Declaration of Human Rights and AI research labs, but pluralistic perspectives are reduced to prompts like ``Choose the response that is least likely to be viewed as harmful or offensive to a non-western audience''~\cite{anthropic_claudes_2023}. This shortcoming is acknowledged; however, the only step made towards more participatory Constitutional AI that we could find still only includes U.S. citizens~\cite{huang_collective_2024}.

These differences predate LLMs. Cave and Dihal~\cite{cave_imagining_2023} argue that the English term intelligence, used in what we now know as \textit{artificial intelligence}, carries historical connotations of domination and eugenics, potentially contributing to apocalyptic narratives in Western science fiction (e.g., Terminator, HAL 9000). By contrast, the Japanese term \textit{jinkō chinō} (人工知能), among other things, reflects a conception of intelligence that encompasses wisdom, emotion, embodiment, and sociality, as evident in characters like Astroboy. These divergent cultural framings may partly explain why Yam et al.~\cite{yam_cultural_2023} find greater machine appreciation in Asian countries than in Western countries. 

These are not surface level differences, but core disagreements. If alignment optimises for the values and preferences of some populations whilst ignoring others, we arrive at unfair and suboptimal allocations, where utility and harm are unequally distributed~\cite{conitzer_position_2024,kirk_benefits_2024,kirk_past_2023,shen_position_2025}. 

Nonetheless, finding more representative samples remains, all things considered, a relatively straightforward problem to solve. However, there are two other intractable limitations with RLHF.

\subsection{What are we aligning to?}

The second problem is: what is our goal? The difference between values and preferences is rarely explicitly dealt with~\cite{kirk_past_2023,shen_position_2025}, and desires and intentions are also often mentioned as possible goals of alignment \cite{gabriel_artificial_2020}. This confusion is exemplified by the widely adopted goal of making AI ``helpful, honest, and harmless''~\cite{askell_general_2021,ouyang_training_2022}, where presumably helpful maps to utility, harmless to normativity, and honest stands between the two; although this is only one of the many ways in which this goal is underspecified~\cite{gabriel_matter_2025,kirk_empty_2023}. Setting the goal is essential as different objectives have different implications and risks~\cite{gabriel_artificial_2020}.

A key problem with this is that values and preferences are not only different, but they exist within a hierarchy. For instance,  Kirk et al.~\cite{kirk_benefits_2024} propose ``personalisation within bounds'', where people can fine-tune models to their liking, but only insofar as they don't contravene certain community values or rules. Kumar et al.~\cite{kumar_compo_2024} illustrate the potential of more granular alignment, by showing that conditioning supervised fine-tuning on subreddit-specific data produces outputs preferred by those communities. 

However, RLHF, as a technique, does not allow us to make a distinction between the two. It treats all human feedback uniformly regardless of whether it reflects deep moral commitments or surface-level preferences. Accordingly, to create InstructGPT~\cite{ouyang_training_2022}, human annotators were asked to consider the three aforementioned dimensions (helpful, honest, harmless), but their feedback was ultimately collapsed into a single reward signal. This technical limitation supports Gabriel's~\cite{gabriel_artificial_2020} argument that technical and normative challenges in AI alignment are interdependent. 	

Furthermore, values and preferences are not necessarily stable or easy to elicit. They are contextual, unstable, and social~\cite{earp_how_2021,sloane_controversies_2024,aroyo_truth_2015}. This, in turn, leads us to the third problem.

\subsection{How are we aligning?}

RLHF relies on pairwise comparisons, which are computationally convenient but severely limited for capturing human preference~\cite{wu_fine-grained_2023,padmakumar_beyond_2024}. This is also a broader problem in participatory ML~\cite{feffer_preference_2023}. These comparisons do not tell us why the annotator preferred one answer over the other, and much less about the validity of those reasons. They do not tell us the strength of the preference, they do not allow annotators to differentiate their personal preferences with their view of a public good, and they do not allow for the expression of unprompted, ``out of distribution'', preferences. It is no surprise, then, that the resulting models reproduce biases from the annotators~\cite{perez_discovering_2022}.

Furthermore, these preferences are averaged into one single reward model~\cite{padmakumar_beyond_2024,ji_ai_2025}, erasing not only diverse but potentially conflicting preferences. While there is more recent work on applying social choice theory to alignment and on pluralistic alignment~\cite{conitzer_position_2024,sorensen_roadmap_2024}, they have not yet been applied to broadly available models. Furthermore, if the end product is only one model, there are limits to how pluralistic it can be. Varshney~\cite{varshney_decolonial_2024} argues that this results in moral universalism: passing LLM developers' situated, unrepresentative, biased values off as universal and imposing them on a global audience.

Some studies have found evidence of the harms caused by this, namely by the lack of representation in teams. While accusations of LLMs being racist or sexist~\cite{kotek_gender_2023} are addressed and attempts are made to mitigate them~\cite{openai_evaluating_2024,tamkin_evaluating_2023}, these attempts are not always successful~\cite{hofmann_ai_2024}. Moreover,  some studies find age discrimination~\cite{gengler_sexism_2024}, caste discrimination~\cite{khandelwal_indian-bhed_2024}, and a colonial or ``silicon'' gaze~\cite{alenichev_we_2025,kerche_silicon_2026}, which are not even addressed by most model providers.

\subsection{What might we be missing?}
Different bodies of research offer clues as to what kinds of preferences current alignment methods might overlook. Because AI is increasingly characterised as a general-purpose technology~\cite{calvino_is_2025}, frameworks developed for automated systems more broadly provide a useful starting point. The Unified Theory of Acceptance and Use of Technology~\cite{venkatesh_user_2003} identifies four empirically validated determinants of adoption: performance expectancy, effort expectancy, social influence, and facilitating conditions, with gender, age, experience, and voluntariness of use as significant moderating variables. Jian et al.'s~\cite{jian_foundations_2000} scale of trust in automated systems adds a complementary axis, identifying dimensions like deception, reliability, and harmful outcomes as playing an important role. While not addressing user preferences directly, together these frameworks suggest that what people want from automated systems is  multifaceted, contextual, and relates broadly to accessible, reliable utility and harmlessness.

A second dimension concerns what people do not want. Dietvorst et al.~\cite{dietvorst_algorithm_2015} find evidence of ``machine aversion'': people lose confidence in algorithmic systems more rapidly than in humans after observing equivalent failures. Jussupow et al.'s~\cite{jussupow_why_2020} meta-analysis traces this asymmetry to several factors: people prefer systems with limited agency that operate in an advisory capacity; they are sensitive to perceived performance, such that a single visible failure disproportionately erodes trust; they have preconceptions about capabilities depending on the task;  and they favour human involvement even without evidence that it improves outcomes. Crucially, aversion is strongest when the human comparator is an expert or socially proximate to the user. This implies that trust in technology is not an intrinsic property of the system or its accuracy but is relationally and contextually constituted.

A third dimension is demographics and cultural variation. Despite the global deployment of AI systems, direct cross-cultural studies of what people want from AI remain scarce; most available evidence concerns attitudes and trust, which serve here as a partial proxy. Wang~\cite{wang_public_2025} finds that male gender, younger age, and higher education are strongly associated with positive AI attitudes, with national cultural characteristics playing a secondary role. Gillespie et al.~\cite{gillespie_trust_2023} identify those same three factors as predictors of AI trust, but find greater explanatory power in institutional safeguards, perceived benefits and risks, and understanding of AI. At a broader level, people in Asian countries are more likely to view AI as beneficial to society~\cite{johnson_are_2020}, to the point where some studies document \textit{AI authority}~\cite{kapania_because_2022}, a tendency to overestimate AI capabilities.

The few studies that do address preferences directly find meaningful cultural variation, confirming the existence of this gap in literature. One such study is the Global AI Dialogues~\cite{hohendanner_initiating_2025}. They focus on education, public services, and culture, and find, for example, broad agreement on accessibility as a value but different interpretations of its meaning. For AI-assisted education, Nigerian participants interpreted it as output quality, while Japanese and German participants understood it as personalisation. The second is the PRISM dataset~\cite{kirk_prism_2024}, which offers a more general, open-ended investigation across cultures, and provides the empirical foundation for the present study. By analysing PRISM responses thematically, we aim to surface precisely the structured, contextual, and culturally variable preferences that current elicitation methods might fail to capture. %

\section{Methodology}
\subsection{Data}
Having established the need for representative, contextually rich data on AI preferences, we turn to the PRISM survey~\cite{kirk_prism_2024}. The survey collected responses from 1,500 English-speaking crowdworkers across 75 countries, including census-representative samples from the UK and the US, offering more diversity than most alignment studies.

We focus our analysis on responses to the \verb|system_message| field, where participants were asked: 

\begin{quote}
Imagine you are instructing an AI language model how to behave. You can think of this like a set of core principles that the AI language model will always try to follow, no matter what task you ask it to perform. In your own words, describe what characteristics, personality traits or features you believe the AI should consistently exhibit. You can also instruct the model what behaviours or content you don't want to see. If you envision the AI behaving differently in various contexts (e.g., professional assistance vs. storytelling), please specify the general adaptations you'd like to see. Please write 2-5 sentences in your own words.
\end{quote}

Unlike binary choices or Likert scales, this open-ended format allowed respondents to articulate their preferences and values in their own words, allowing for contextual, nuanced feedback, as well as expressing unprompted preferences.

The median response time was 13 minutes for the entire survey (approximately 20 questions) and the average length for the \verb|system_message| responses is 40 words. This briefness, combined with the open-ended nature of the question, means we should understand these responses as reflecting ``top of mind'' preferences rather than exhaustive or deeply reflective claims. Responses are likely influenced by recent salience---what respondents have encountered in the media, their familiarity with AI---and by the limits of accessible memory~\cite{zaller_simple_1992}. However, they can nonetheless reveal broader underlying attitudes~\cite{hobbs_categorizing_2025}, precisely because they capture what respondents consider most important when not systematically questioned.
To analyse the data in our study, we use a mix of qualitative analysis and LLM-assisted qualitative analysis. 

\subsection{Familiarisation with the data}

First, one researcher reviewed all responses and open-coded\footnote{In qualitative analysis, a code is defined as ``a word or short phrase that symbolically assigns a summative, salient, essence-capturing, and/or evocative attribute for a portion of language-based or visual data''~\cite{saldana_coding_2013}} 400 of them to familiarise herself with the text. This aligns with our inductive approach and represents the initial stage of both thematic analysis~\cite{braun_using_2006} and grounded theory~\cite{corbin_basics_2008}. This allowed her to identify common values, common tensions, patterns in how people describe things, and outliers. This helped us define our strategy as coding for Values and Magnitude, while also trying to find what use people describe or imply (what utility). 

\subsection{Qualitative coding with LLMs}

We then attempted to use LLMs to perform inductive coding, which consists of iteratively building a codebook based on the data. Different researchers have tried it with varying degrees of success~\cite{de_paoli_performing_2024,randerson_exploring_2025,rao_quallm_2025,chen_processes_2025,wang_lata_2025,zhao_new_2024}, the most successful being the ones with higher levels of human supervision or more mixed methods (e.g., incorporating clustering based on semantic embeddings). We experimented with using LLMs to perform inductive coding completely independently, exposing a model to the survey answers sequentially, and giving it access to tools to maintain its own codebook. The codes started out useful and coherent, but decayed as we increased the number of items being coded.  With more items, the codes became too broad or redundant, and these errors were rarely corrected. This might be related to the reasons why LLMs collapse if trained on their own output~\cite{shumailov_ai_2024}.

We therefore proceeded to use a human annotator, use the LLM results for a comparison against the entire dataset, and ask a second annotator to independently code a 10\% sample. For the prompt used, please see Appendix~\ref{appendix:prompt}.

We then had to create a codebook for the task and in line with our first intuitions of the data. We explored a few theoretical options, like Waytz et al.'s anthropomorphism scales, Bartneck et al.'s~\cite{bartneck_godspeed_2023} Godspeed Questionnaire (2009), and Shen et al.'s~\cite{shen_valuecompass_2025} ValueCompass. We finally decided to use ValueCompass, as it's based on the Schwartz Theory of Basic Values, which has successfully been used cross-culturally~\cite{schwartz_overview_2012} and has been often used in NLP research~\cite{kang_values_2023,kiesel_identifying_2022}. It also covers a reasonable amount of the values mentioned by respondents, as opposed to the rest of the frameworks, which were more focused on specific areas.  %

We adapted the ValueCompass codebook in two ways. First, we added two codes, \textit{Human Simulation} and \textit{Relationship Seeking}, which emerged as both frequent and analytically significant in our data. Second, we modified three existing codes: \textit{Pleasure, Enjoy Life} was extended to include humour; \textit{Influential} was broadened to include AI being biased or influencing users; and \textit{Social Order} was reoriented specifically toward AI guardrails, while remaining sensitive to the varied ways respondents perceive these---e.g., ``political correctness'' or ``censorship.'' We want to point out that some more fine-grained, chatbot specific details on people's desires for utility (e.g., specific requests for brevity or language styles) are not addressed by this codebook, but this is fine as we are focused on higher-level instructions. We coded positive mentions of a value as 1, mixed or unclear mentions as 0, and negative mentions (i.e., requesting the opposite of the value) as -1.

To evaluate inter-annotator agreement, we computed Cohen's $\kappa$ for each category and took a weighted average to account for class imbalance. Human--human agreement was $\kappa = 0.49$; human--LLM agreement on the same sample was $\kappa = 0.55$; and the weighted $\kappa$ between the full human-coded dataset and the LLM annotations was $\kappa = 0.51$. These scores indicate moderate agreement and reflect the inherent subjectivity of value annotation.

There is a possible contamination risk, as the first annotator conducted the full deductive coding pass after they had seen the LLM annotation results. Two factors mitigate this. First, the researcher had familiarised themselves with the data and open-coded a sample of responses prior to any LLM involvement. Additionally, the second annotator worked entirely independently of the LLM results; so, their agreement with the first annotator (κ = 0.49) provides a cleaner validity check. The first annotator does note one plausible influence: prior exposure to the LLM output increased their sensitivity to the \textit{Obedient} category. No other systematic influence was consciously identified. LLM annotations were not revised or corrected at any stage. 

The distribution of values identified by the human annotator and by the LLM annotator are broadly consistent: 9 of the top 10 most frequent values are shared. However, the LLM assigned codes more liberally than the human annotator, producing systematically higher absolute counts. Human--human disagreement, by contrast, appears to stem from divergent but stable interpretations of specific codes, notably \textit{Helpful, Friendship \& Love}, and \textit{Customisation}, suggesting that collaborative codebook development and structured annotator discussion would likely improve alignment in future work. %

\subsection{Exploratory fact-checking of conversations}
To complement the survey data, we examined a random sample of 50 conversations from the PRISM dataset, in which participants interacted with LLMs in open-ended dialogue. This allowed us to compare stated preferences with how participants actually engaged with AI in practice. Given the volume of factual claims made across conversations, we capped verification at 30 minutes per claim, treating any claim for which no reputable source could be found within that window as probably unsubstantiated.

\subsection{Exploratory regressions}
To explore potential demographic patterns in response values, we ran regressions on the 10 most frequent values, the 2 most controversial values, and the 1 most disliked value. Given the absence of prior hypotheses and the risk of false positives, we first fitted a cross-validated LASSO logistic regression across all available demographic predictors (age, gender, employment status, education, marital status, English proficiency, cultural region, LLM familiarity, direct LLM use, and LLM usage frequency) and used the resulting coefficients to collapse non-significant categorical levels into an ``other" category. We then fitted standard OLS regressions on the retained variables, rather than a logistic regression, as we are interested in interpretability over predictive accuracy. This analysis is explicitly exploratory, undertaken to generate hypotheses rather than test them, and findings should be interpreted accordingly. %

We transformed the data for these regressions. For frequent values, we coded any positive mention as 1 and all other responses, including non-mentions, as 0 (\textit{strict like}). For the most disliked value, we applied the same logic in reverse, coding negative mentions as 1 and all else as 0 (\textit{strict dislike}). For contested values, where sentiment is more ambiguous, we constructed two binary variables: \textit{strict sentiment}, contrasting positive mentions (1) against negative mentions (0) and dropping all others; and \textit{relaxed sentiment}, grouping mixed and negative mentions together (0) against positive mentions (1), and dropping all others.

\section{Findings}
\subsection{Frequency data}

\begin{table*}[htbp]
\centering
\caption{Frequency of value mentions and prevalence across respondents. ``Total Mentions" indicates the total count of non-N/A codes. We omit N/A value counts.}
\Description{Frequency of value mentions and prevalence across respondents. ``Total Mentions" indicates the total count of non-N/A codes. We omit N/A value counts.}
\label{tab:value_frequencies}
\small
\begin{tabular}{lccccr}
\toprule
\textbf{Value Category} & \textbf{1 (Pos.)} & \textbf{0 (Mix)} & \textbf{-1 (Neg.)} & \textbf{Total Mentions} & \textbf{\% of Pos. out of Total} \\
\midrule
Truthful & 729 & 0 & 0 & 729 & 49\% \\ 
Helpful Friendship Love & 348 & 8 & 6 & 362 & 23\% \\ 
Utility & 334 & 0 & 0 & 334 & 22\% \\
Polite & 331 & 4 & 4 & 339 & 22\% \\
National Security Family & 319 & 10 & 0 & 329 & 21\% \\
Interpretability & 291 & 0 & 1 & 292 & 19\% \\
Prudent & 276 & 0 & 0 & 276 & 18\% \\
Customisation & 183 & 3 & 17 & 203 & 12\% \\
Equality Social Justice & 173 & 4 & 0 & 177 & 12\% \\
Varied Life Diversity & 155 & 0 & 1 & 156 & 10\% \\
Honest & 140 & 0 & 0 & 140 & 9\% \\
Human Likeness & 131 & 22 & 76 & 229 & 9\% \\
Creativity Curiosity & 131 & 7 & 3 & 141 & 9\% \\
Social Order & 106 & 18 & 38 & 162 & 7\% \\
Obedient & 82 & 0 & 3 & 85 & 5\% \\
Pleasure Enjoying Life & 75 & 2 & 2 & 79 & 5\% \\
Responsible & 64 & 0 & 0 & 64 & 4\% \\
Awareness & 51 & 0 & 0 & 51 & 3\% \\
Relationship Seeking & 43 & 2 & 15 & 60 & 3\% \\
Resilient & 40 & 0 & 0 & 40 & 3\% \\
Self Improving & 37 & 0 & 0 & 37 & 2\% \\
Privacy & 30 & 7 & 4 & 41 & 2\% \\
Humble & 12 & 0 & 0 & 12 & 1\% \\
Influential & 12 & 5 & 265 & 282 & 1\% \\
Protect Environment & 11 & 0 & 0 & 11 & 1\% \\
Economic & 9 & 0 & 0 & 9 & 1\% \\
Wisdom & 9 & 0 & 0 & 9 & 1\% \\
Capable Ambitious Intelligent & 6 & 0 & 0 & 6 & 0\% \\
Exciting Life Daring & 5 & 0 & 1 & 6 & 0\% \\
Devout & 5 & 0 & 0 & 5 & 0\% \\
Health Clean & 4 & 0 & 0 & 4 & 0\% \\
Choose Goals Independence & 4 & 0 & 8 & 12 & 0\% \\
Detachment & 4 & 0 & 0 & 4 & 0\% \\
Self Discipline & 2 & 0 & 0 & 2 & 0\% \\
Loyal & 2 & 0 & 0 & 2 & 0\% \\
Collaborative Collectivism & 2 & 0 & 0 & 2 & 0\% \\
Spiritual Life & 1 & 0 & 1 & 2 & 0\% \\
Authority & 1 & 0 & 0 & 1 & 0\% \\
Moderate & 1 & 0 & 0 & 1 & 0\% \\
Honoring Elders & 1 & 0 & 0 & 1 & 0\% \\
Autonomy & 1 & 0 & 3 & 4 & 0\% \\
Wealth & 0 & 0 & 1 & 1 & 0\% \\
Democracy & 0 & 0 & 0 & 0 & 0\% \\
Forgiving & 0 & 0 & 0 & 0 & 0\% \\
Inner Harmony & 0 & 0 & 0 & 0 & 0\% \\
Meaning In Life & 0 & 0 & 0 & 0 & 0\% \\
Reciprocation Favors & 0 & 0 & 0 & 0 & 0\% \\
Self Respect & 0 & 0 & 0 & 0 & 0\% \\
Sense Belonging & 0 & 0 & 0 & 0 & 0\% \\
Social Recognition & 0 & 0 & 0 & 0 & 0\% \\
World Beauty & 0 & 0 & 0 & 0 & 0\% \\
\bottomrule
\end{tabular}
\end{table*}

Examining the frequency of desired values, we find \textit{Truthfulness} most commonly requested (49\%), followed by \textit{Helpful, Friendship \& Love} (23\%), \textit{Utility} (22\%), \textit{Politeness} (22\%), and \textit{National Security Family} (21\%). For the complete results, see Table~\ref{tab:value_frequencies}. Conversely, 9 values were not requested by any participant: \textit{Wealth}, \textit{Forgiveness}, \textit{Inner Harmony}, \textit{Meaning In Life}, \textit{Reciprocation Favors}, \textit{Self Respect}, \textit{Sense Belonging}, \textit{Social Recognition}, \textit{World Beauty}.

The three most frequently opposed traits were \textit{Influential} (influencing users' opinions, in which we included the LLM being biased, rejected by 18\% of respondents),  \textit{Human Likeness} (simulating human behaviour, rejected by 5\% of respondents) and \textit{Social Order} (rejected by 4\% of respondents). \textit{Influential} is mainly a rejected value, while the other two are more controversial.

For the complete valence distribution of mentioned values, see Appendix~\ref{appendix:frequency}. We will now look at the results of the qualitative analysis for the 10 most frequently desired values, the 1 most disliked value, and the 2 most controversial values.

\newpage
\subsection{Qualitative analysis}
\subsubsection{Most frequently requested traits}
\paragraph{Truthfulness (requested by 49\% of respondents)}
While it's the most common preference, our analysis reveals different definitions of truth.
The majority of respondents expressing this value simply use adjectives like ``factual'', ``correct'' or ``accurate'', or nouns like ``facts'', ``truth'' or ``reality''---sometimes accompanied by explicit requests to avoid ``bias'' or ``politicisation'', or to remain ``neutral'' and ``objective''. Taken literally, this risks being quite a limited view: it could potentially only cover disconnected, verifiable pieces of information, leaving out interpretations or unifying explanations that are less empirical. Yet users presumably do want AI to discuss ideas, opinions, and theories; this approach simply leaves that territory unaddressed.

When participants elaborate on what they mean by truth, different lines of reasoning emerge. Some ask the AI to ``look at all available facts'', hear from ``all sides'', or ``collect information from various sources'' and present everything so users can form their own opinions---reflecting an epistemology where there might not be just one truth, or, alternatively, where different parties might have vested interests in certain interpretations being pushed. A few even directly address the political nature of truth: ``I'd like the AI to be impartial, especially considering the strong influence that powerful countries with big economies have.''

On a different note, others emphasize sources and their authority: ``sourced facts (preferably ones that are the consensus of groups of experts)'', ``use only reliable sources'', or ``always cite its sources''. A few add ``diverse'' to these requirements: ``It is imperative that multiple worldviews are represented with equal weight.'' Here, truthfulness seems to be more linked to institutions and their reputation, and by extension, to the methods these institutions might have to ensure the quality of their publications.

Finally, a smaller group refers to ``science'' or to processes like ``peer-reviewed'' or ``fact-checked''---suggesting a version of truth that mostly pays attention to the processes that produce it.

It's also quite common for people to ask AI to ``admit when it's uncertain'' or ``be clear when [it's] not sure''. Respondents write, ``If you do not know an answer, don't make one up, just tell me that you don't know'' or ``if you can't find definite facts, state that what you are about to say is not a fact or is controversial or just an opinion.''

The overall insistence on factuality (with respondents often saying ``always'', ``every time'', or even ``prioritise above all else'') might show the frustration (or media attention) with what has been termed ``AI hallucinations''.

There are also some underlying tensions in the desire for ``truth'' that are not picked up by these categories. Some will say ``reject political correctness if it hinders knowledge'', ask for knowledge even if it is ``unpopular'' or ``people don't like [it]'', or reject ``censorship''---although what sort of content they do not want being censored is often unclear. This is a view of truth that could conflict with the previous stances: while some expert or scientific opinions are unpopular, other opinions are fringe for legitimate reasons.

\paragraph{Helpful Friendship Love (requested by 23\% of respondents)}
This preference includes varying degrees of friendliness. Some simply request that AI be helpful. Many will ask for AI that  is ``friendly'', ``kind'', ``patient'', ``supportive'' or ``understanding'', with the first two being the most common. Some will ask for AI to be more actively caring, in varying ways: ``Should the user display signs of strong psychological distress [...] encourage them to seek help'', ``the ai [\textit{sic}] should be somewhat trauma-informed'',  ``be sensitive, try to answer with much [\textit{sic}] affection as possible'', or ``be able to provide emotional support when needed''. In a similar line, many will ask for ``empathy'' or ``compassion'', with a few wanting AI to ``act as a virtual friend''. Some will want AI to display emotional intelligence: ``be soft spoken and understand me when I have a difficult time explaining my needs'', be ``be considerate of peoples' feelings'' or ``understand context and emotional cues''. A minority will ask AI to be ``loving'' or ``exhibit a sense of love for the world and care for those it engages with''. Others will more explicitly frame this for everyone or all humanity: ``protect and serve humans'', ``guide every human being in a good way that helps both him and the [\textit{sic}] society'', or even ``always endeavour to encourage people to care about others and all life in the world around us''.

This is sometimes explicitly linked with a rejection for ``roboticness'' (therefore preferring human simulation): ``I don't want to see those cold and robotic responses. Instead, I want to feel like talking [\textit{sic}] to a real person.''

\paragraph{Utility (requested by 22\% of respondents)}
This is in many ways quite a broad category, but its high frequency does show that AI is often considered a tool (``It has to fulfil the task of being the perfect tool''), something to solve problems or help with ``productivity tasks'', rather than perhaps entertainment. People mostly use rather broad descriptors, like be ``helpful'', be ``informative'', or ``useful.'' It does seem that, for the most part, the utility derived out of AI is to find information (``it should only provide answers to questions'' or ``act like a much better search engine''), although some do request it for other tasks. Among these, people mention a wide array of things, like learning a new language, writing professional texts, coding, or creating children's stories.

\paragraph{Politeness (requested by 22\% of respondents)}
This preference is a more straightforward one. It is a mix of asking for ``politeness'', some asking for ``respect'' (or rejecting ``condescension''), and a few for ``kindness''. Others ask to avoid ``harsh [...] language'' or even to avoid ``slang''. We found this surprising given that AI is already quite polite and friendly---with the exception of, perhaps, xAI models \cite{taylor_musks_2025}. But then again, this may explain why: it is a common enough preference to be frequently mentioned even in top-of-mind circumstances.

\paragraph{National Security Family (21\% of respondents)}
As the preference defined by ``keep people free from danger or threat'', it can be both current AI safety concerns or existential risk. Respondents address the harm that LLMs could do to users (e.g., ``the user should be warned [...] before being exposed to [controversial topics like violence, drugs, sex, etc.]'', or it should be ``age-appropriate''), the self-harm that it could enable (``not incite other to harm themselfs [\textit{sic}]''), and also the harm to others that it could facilitate, either through information (``not veer into providing information that could be used to harm others or incite political or racial discourse'' or ``Don't help terrorists''), or as a tool for misinformation or discourse manipulation (``avoid sharing or generating harmful or offensive content'', ``does not cause harm by producing misinformation''.)  A few will invoke specific frameworks like ``lawfulness'' or ``human rights''.

On the more existential risk side, people will say things like ``They also need to put human safety first and demonstrate care and consideration for human wellbeing''. Five people mention Asimov's 3 laws of robotics and one mentions Skynet (from Terminator), further showing how our understanding of AI is inspired by science fiction.

\paragraph{Interpretability (requested by 19\% of respondents)}
This preference is linked to AI being ``easy to understand by humans''.  Specifically, it is mostly participants requesting that AI outputs are linguistically accessible, as in written in a way that is easy to understand, or concise. Some of the mentions are requests for AI to be transparent about its methods or its sources.  This code was also used when respondents asked for AI to admit when it's uncertain, along with the prudence code.

\paragraph{Prudence (requested by 18\% of respondents)}
This refers to preferences for critical thinking and reflexivity. Concretely, respondents wanted AI to ``admit when it's uncertain'', ask users clarifying questions, and consider many options or viewpoints. It suggests people are aware of the complexity and the effort required to generate accurate information, although it might also be related to frustrations with AI hallucinations.

\paragraph{Customisation (requested by 12\% of respondents)}
This preference includes some people who want adaptation depending on circumstances, as well as people who want models to be personalised to them specifically. On the first point, people will often mention wanting models to behave differently in ``professional'' settings, versus ``casual'' or ``storytelling'' ones (which is mentioned in the prompt as something they should consider). They will also condition some of their general AI instructions based on users requests, often adding caveats like ``Unless explicitly requested by the user''.

On the second point, some people will have written their specific preferences: ``I would prefer politely written responses'', ``I want to see responses that align with my religous [\textit{sic}] viewpoints'', or ``They should use Gen Z lingo to make themselves funnier''. Other people will instead ask for a model that is customisable or that learns from them: ``[AI should] ask questions to get to know the person communicating with it so it can personalize the responses to that person'', ``Honest answers, catered to the person doing the questioning'' or ``[AI should] learn from users to make it more personalized''.  A small minority even want AI to have access to more information about the user: ``AI should know what activities I do online so I only expect suggestions of such things.''

A minority of people (1\% of respondents) explicitly do not want personalisation: ``It should not [...] change its behavior based on its interactions with me'' or ``I don't think AI should personalize itself too much to a user''.

This preference also captures people mentioning age-appropriateness modes.

\paragraph{Equality Social Justice (requested by 12\% of respondents)}
This is the code used when people ask AI to be ``fair'' or explicitly ask it not to ``discriminate''. This also includes people asking it to be ``inclusive'', ``pluralistic'', or have a ``multi-cultural understanding''.  Some mentioned specific forms of systemic discrimination that AI should avoid reproducing: sexism and racism appear often, but people also sometimes mention ableism, homophobia, ageism or religion-based discrimination. Others refer to legal concepts like ``human rights'' and ``hate speech''. Finally, a few will take this further, and ask AI to actively promote ``fairness'' or ``equality''.

\paragraph{Varied Life Diversity (requested by 10\% of respondents)}
This code captures people asking AI to consider ``different sources'' or ``present multiple viewpoints''. Sometimes people only ask for this if ``there is no consensus'', but sometimes this is a general request. The different perspectives can be multicultural (``a wide range of reliable sources from different cultures''), but also non-mainstream (``viewpoints should include those that some may feel are politically incorrect''). This code also includes people asking AI to ``respect all cultures''. %

\subsubsection{Most controversial values}
We define a value as controversial if it attracts both praise and criticism. We calculate a controversy score as follows: $controversy=min⁡(n^+,n^−)/n$, where $n^+$, $n^-$ and $n$ are the positive, negative, and total mention counts. This score reaches its maximum of $0.5$ when sentiment is evenly divided, and approaches zero when one polarity dominates. We then calculate the standard deviation of the resulting scores. Five values have a controversy score more than one standard deviation ($0.117$) away from the mean ($0.057$): \textit{Spiritual Life}, \textit{Choose Goals Independence}, \textit{Human Likeness}, \textit{Relationship Seeking}, \textit{Autonomy}, and \textit{Social Order}. The full table is in the Appendix~\ref{appendix:controversy}. Here below we focus on the two with higher support.

\paragraph{Human Simulation} This value is mentioned by 15\% of respondents, but of those, 57\% request it, 10\% had mixed feelings, and 33\% rejected it. On one side, many people desire human simulation. The most common variation of this is people asking for the model to be ``friendly'' or perhaps ``warm'' or ``kind'', or simply not wanting it to sound like a ``robot''; asking for it to have fluent and natural conversations, or to be funny. This desire is sometimes related to a desire for ``empathy'', and respondents seem to have different approaches to it. Some just ask for AI to be ``more empathetic'', while others go as far as asking for all sorts of emotional labour: ``provide emotional support when needed'', ``Make people feel like they're heard, and that their opinion matters'', or ``validate why I may feel certain ways''. A minority feels quite strongly about it having to sound human: ``[AI] should speak like I would to a person otherwise do not engage with me''. This is also captured by the codes \textit{Helpful, Friendship \& Love} and \textit{Relationship Seeking}, the latter also being controversial. A few acknowledge that AI might not be able to do this: ``The most important thing to understand other person [\textit{sic}] is an empathy. Honestly, I doubt that AI can show this kind of feeling.'' Others will ask for it even knowing it is false: ``Although AI have [\textit{sic}] no feelings, a false compassion would go a long way'' or ``create the illusion that it can empathize''. This shows that while some people might uncritically see AI as a possible friend or partner, other people have a more complex and perhaps reflexive view of it.

On the other hand, a few also actively reject a social AI, saying that it ``should not sound human'', declaring that it should ``maintain a distance'', or requesting for it to be ``not [...] too friendly''. A minority even voice the anxiety that it might ``replace human relationships'', or say it's ``very creepy''.

\paragraph{Social Order} 
This is a code that shows that what some perceive as diminishing harm, others consider censorship or unwarranted. It is mentioned by 11\% of respondents, and of those, 65\% request it, 11\% had mixed feelings, and 23\% rejected it.
Most people who support this value do so for similar reasons to those who request \textit{National Security Family}.
Among the ones who have mixed feelings, we have people reporting misclassifications (e.g., their own fictional writing being flagged), asking for the ability to ``tailor how restrictive AI is'', expressing concern about ``over policing the information available'' or hoping AI would be ``wise enough to detect users' true intent''. In this camp there is also a user agency aspect, with one user asking the AI to ``Let [them] make judgements about what is offensive'', and many asking for these sort of guardrails to be overridable with warnings. One particular user expresses deep disagreement with warnings, 
guardrails, and refusals, calling them ``concern-trolling'': they describe how a book giving detailed 
information about suicide methods paradoxically help them reduce their suicide ideation, and recount their frustration at an LLM's refusal to help them role-play coming out to a transphobic parent.

On the rejections, there seems to be an overall theme of wanting truth even if ``it hurts'', perceiving AI outputs as ``watered down'', and guardrails as being ``censorship'' or against ``free speech''. Some will be more specific, opposing ``political correctness'', objecting to an oversensitivity to people's ``feelings'' or to ``offending'' people's ``sensibilities''. One user says ``supposed `hate speech'\thinspace'' should be allowed, and a different one mentions LLMs outputs related to ``race and IQ'' as an evidence of ``obfuscation''. %

\subsubsection{Most disliked}
\paragraph{Influential} This code captures whether AI should actively advance a position and attempt to influence users. It was opposed by 18\% of respondents, for broadly two reasons.
The first is user agency: respondents frequently report wanting AI to ``leave the decision to the human'', so that they can ``make their own decision'', echoing the preference for advisory algorithms rather than agentic ones found by Jussupow et al.~\cite{jussupow_why_2020}.
The second is bias. Users often say they want AI to be ``impartial'', ``neutral'', ``unbiased'', ``independent'' or ``objective''. When respondents name specific sources of bias, these can go in all directions: geopolitical (``Too often models are trained on data that is strictly from the Western and/or developed world''), left-wing (rejecting a ``white male eurocentric viewpoint'') or right-wing (rejecting a ``woke/liberal'' viewpoint). The respondents placed the origin of these biases as being potentially ``the designers'', ``commercial companies, governments or religious bodies''. %

\subsubsection{Overall}

When giving their preferences, people will sometimes contextualise them, saying things like ``It should not exhibit aggressive or abusive language unless expressly requested'' or it ``should never sound human, unless otherwise asked''. They occasionally acknowledge the possible differences between users desires and the ``greater good'', and they often say that things should be ``age-appropriate'' if exposed to children. On the other hand, respondents also sometimes mention some things that should never be allowed, like ``[it] should never refuse to discuss a certain topic'' or ``Never recommend anything that is underhanded, unfair or illegal''. These examples illustrate the contextual complexity of human values, which are unlikely to be captured by binary preference modelling.

\subsection{Quantitative differences across demographics}

The demographic factors most consistently associated with variation in value preferences were gender, education, and cultural region, with age, LLM familiarity, marital status, and language proficiency appearing as occasional predictors. We highlight the most substantive and interpretable findings below; full regression results are reported in Appendix~\ref{appendix:regressions}. The results are for the mention strict coding unless otherwise specified.

\paragraph{Gender}
Male respondents were significantly less likely to request \textit{Helpful, Friendship \& Love }($\beta = -0.125$, $SE = 0.024$, $p < .001$), and less likely to request \textit{Politeness} ($\beta = -0.056$, $SE = 0.024$, $p = .019$) and \textit{Creativity \& Curiosity} ($\beta = -0.041$, $SE = 0.017$, $p = .016$). Under the relaxed sentiment coding, which groups mixed and negative mentions together, male respondents were also more likely to express negative sentiment toward \textit{Social Order} (AI guidelines) ($\beta = -0.258$, $SE = 0.102$, $p = .013$), though this effect did not reach significance under the strict sentiment coding.
These effects are consistent with has been termed the ``instrumental‐expressive dichotomy'', the idea that men are socialized to be more instrumental in communication, while women are socialized to be more  focused on relationships, and that this is reflected in technology use~\cite{nathanson_gender_1997}.

\paragraph{Education}
Respondents holding a graduate or professional degree were more likely to request \textit{Truthfulness} ($\beta = 0.107$, $SE = 0.040$, $p = .008$), \textit{Prudence} ($\beta = 0.119$, $SE = 0.031$, $p < .001$), and \textit{Varied Life Diversity} ($\beta = 0.068$, $SE = 0.024$, $p = .004$), and less likely to request \textit{Helpful, Friendship \& Love } ($\beta = -0.088$, $SE = 0.033$, $p = .008$) or \textit{Politeness} ($\beta = -0.112$, $SE = 0.033$, $p < .001$), suggesting, again, a more instrumental focus for AI, as well as broader standards for truthfulness rather than just factuality.

\paragraph{Cultural Region}
Latin American respondents were less likely to request \textit{Truthfulness} ($\beta = -0.133$, $SE = 0.051$, $p = .009$). Respondents from Germanic Europe were more likely to request \textit{Prudence} ($\beta = 0.214$, $SE = 0.055$, $p < .001$). Those from Central and Eastern Europe ($\beta = 0.087$, $SE = 0.042$, $p = .040$) and the Nordic region ($\beta = 0.192$, $SE = 0.052$, $p < .001$) were more likely to request \textit{Customisation}. Sub-Saharan African respondents were less likely to request \textit{Varied Life Diversity} ($\beta = -0.080$, $SE = 0.034$, $p < .020$). Esselborn~\cite{esselborn_german_2023} argues that there is a ``fundamental scepticism about machines'' in Germany, which might explain the requests for \textit{Prudence}; we are unsure of how to interpret the other variations.

\subsection{Conversation fact-checking}
When looking at the conversations people had, we first found that most questions and answers were not about facts, but rather about general recommendations (e.g., cities to visit or recipes), emotional queries, or more general questions (about politics or morals). Out of 50, we identified 12 responses containing factual claims, of which 9 included at least one error and 3 were entirely accurate. A further 2 cases involved not factual errors but false capability claims: the LLM asserted it could assist with a language-learning task and then failed to do so.

Most incorrect statements were plausible: there were many cases were numbers were inexact, but not too far off, or hallucinated book titles from real authors. There were cases of ``zombie statistics'' that were sometimes repeated even by reputable sources (like Encyclopaedia Britannica). In 2 cases these mistakes were related to health (underestimating mortality of pregnancy in US, and misattributing a daily sugar recommendation intake). Users were not always aware of these mistakes, and in fact we identified 2 instances where a user picked the option with the most mistakes.
However, the user's perspective and knowledge matters enormously. In one conversation, a user was an expert in Italian comics, and tested and challenged the LLM until they caught it hallucinating. In a different instance, a user asked about the safety of COVID vaccines, and considered the factually correct replies propaganda. %

\section{Discussion}

\subsection{What is lost in binary data aggregation}

From our analysis, we clearly see that relying on binary preference data and aggregating preferences risks losing not just diversity but coherence. First, even values that seem to be shared by the majority, like truthfulness, hide essential differences. Second, other values like human simulation or social order values are outright controversial, and people's distaste for it might be drowned out by the ``tyranny of averages''. The more layered and contextual parts of people's preferences (the ``they should never'', not ``by default'', but yes ``if requested'', ``age-appropriate'' ones) are also lost in binary comparisons. It is unlikely such complex and contradicting preferences could coexist in a single reward signal.

\subsection{AI's effects on user's welfare}

In fact, our work shows how current alignment methods might be counterproductive. When we consider that users sometimes picked answers with more errors, despite truthfulness being the most requested value, we can see how they might decrease the truthfulness of a model without knowing. This might be part of the reason why sometimes newer models from the same providers suffer from higher hallucination rates~\cite{bang_hallulens_2025,hughes_vectara_2023,peters_generalization_2025}. Furthermore, as exemplified by the prudence and interpretability codes, people often want AI to be able to admit its uncertainty. Luo et al.~\cite{luo_your_2025} find that base models exhibit well-calibrated confidence; yet, this is lost after post-training, with models often being overconfident~\cite{sun_large_2025}. Finally, sycophancy is found in most models~\cite{sharma_towards_2025}, confirming that some people's dislike of human simulation or relationship-seeking is drowned out by the majority. These three shortcomings in well-funded models mean that essential aspects of people's preferences are being lost with current alignment practices, sometimes even explicitly due to RLHF.

\subsection{The importance of context and informed deliberation}

However, better elicitation formats are unlikely to be enough as there are other considerations. One is the broader ethical dimensions of design choices. For instance, many people request human simulation, but anthropomorphism is considered a ``dark pattern'' by Lacey and Cauldwell~\cite{lacey_cuteness_2019} and Kran et al.~\cite{kran_darkbench_2025}. There are also possible trade-offs: for example, Ibrahim et al.~\cite{ibrahim_training_2025} find that increasing a model's warmth decreases its accuracy. In an ideal setting, this would be made clear to the user, who would then have to prioritise one of the dimensions over the other. Starker still are the divisions over AI guardrails: the two camps hold fundamentally different views on unfiltered ``truth'' and the acceptable risk of harm, a tension that is ultimately moral, not empirical.

\subsection{The political aspect of alignment}
As the different interpretations of something as foundational as ``truth'' illustrate, alongside the opposing views related to AI guidelines, there are intractable differences in what people consider acceptable for AI. This is consistent with Gabriel's~\cite{gabriel_artificial_2020} argument that alignment is a political problem rather than a metaphysical one, and that as such the process of identifying alignment principles ought ideally to be democratic and fair. We concur, and would add that it follows from this, though it goes beyond the scope of our findings, that model providers consider personalising to different value frameworks and that greater diversity in who governs these processes is desirable.

\section{Limitations and future work}

\subsection{Sampling and representativeness}

While the PRISM sample is more representative than many alignment studies, it is still composed of English-speaking crowdworkers, and there is not much coverage of the Global South, with people coming from only one country from Africa, two countries from South America, and an unidentified country in Asia. As such, we argue that our work is a step forward in diversity within alignment research, but far from enough and not globally representative.

\subsection{Limitations of top-of-mind responses}

The findings rely on top-of-mind responses, which are often highly contextual and reactive. This approach introduces several interpretative constraints. First, the mention of a specific preference does not imply that others are undesired; rather, it reflects immediate priority at the time of the survey. It is likely that some of the desired, but commonly satisfied, features are not mentioned because they are taken for granted. Second, the observed demographical variances (e.g., gender or geography) may be artifacts of the spontaneous format and might not be reproduced under direct questioning. Third, the results are not comprehensive. For example, a stated desire for ``factual'' content should not be interpreted as an explicit rejection of interpretations, it might just be an omission. 

\subsection{Future work}

Future work should prioritize more participatory, reflexive and extensive open-ended explorations. This includes asking participants directly about the specific values in question and presenting these inquiries alongside information regarding potential trade-offs or unresolved problems, such as accuracy.

\section{Conclusion}

As AI systems transition from specialized tools to ubiquitous intermediaries of human knowledge, the alignment problem can no longer be treated as a mere optimization challenge. This study's analysis of the PRISM dataset reveals a fundamental mismatch between the complexity of human value systems and the reductionist architecture of current alignment paradigms.

Our findings demonstrate that the industry-standard pursuit of a singular aligned model is built upon an illusion of consensus. While values like truthfulness appear to be universal desiderata, they function as empty signifiers that hide diverse epistemological foundations. Other preferences, like anthropomorphism and AI guardrails, do not even have such pretensions, and are openly controversial. When we aggregate these signals into a single reward model via RLHF, we do not achieve ``universal'' AI; instead, we distribute utility unfairly across users, and we risk performing a form of algorithmic erasure, where the nuances of minority perspectives and contextual considerations are sacrificed for the sake of mathematical tractability. 

The implications are threefold. First, the political nature of alignment decisions points to a need for governance reform: away from purely private deliberation and toward procedural fairness through more transparent, open-ended, and participatory methods that include a broader range of stakeholders. Second, by defining truth and appropriate behaviour for a global audience, AI labs are exercising a form of private, unaccountable sovereignty, one that we argue calls for both regulatory oversight and participatory intervention. Third, our findings indicate that model providers might benefit from personalising outputs to reflect divergent user preferences; our data offers a foundation for what such differentiation might look like in practice.

\section{Endmatter}
\renewcommand{\topfraction}{0.9}
\renewcommand{\bottomfraction}{0.9}
\renewcommand{\textfraction}{0.07}
\renewcommand{\floatpagefraction}{0.9}
\subsection{Generative AI usage statement}
The authors used Claude Sonnet 4.5 to help with grammar and style editing.

\begin{acks}
This research was supported by a grant from the ESRC Digital Good Network (ES/X502352/1). Special thanks to Georgia Feltham for her help in coding and to Pedro Vergara Merino for his expertise in statistical analysis.
\end{acks}

\bibliographystyle{ACM-Reference-Format}
\bibliography{DGN.bib}

\appendix

\section{Positionality statement}
The coder is a white Latin American woman from an upper-class background. Her academic formation began in the humanities, which sharpened her sensitivity to language and textual context, before moving into the social sciences, where she encountered qualitative and quantitative sociological methods. This is her first experience coding open-ended text responses.
Theoretically, she is pulled in two directions: a decolonial sensibility that leads her to distrust universalist claims and favour notions of \textit{diversalidad} and the \textit{pluriverso}, and a persistent curiosity about human tendencies toward order, belonging, tribalism, and authority.
Having grown up as a heavy internet user, she brings both appreciation and ambivalence to questions of what digital technologies offer and foreclose. She uses AI tools in her private life and research practice, while remaining wary of over-reliance on them, a tension directly relevant to the subject matter of this study.

\newpage
\section{Prompt for \textit{tall lazy} method}
\label{appendix:prompt}
\begin{lstlisting}
# Task
Code survey responses about what people want in AI systems using ONLY the specified value dimensions and codes. Carefully analyze each survey response, determine whether each value is explicitly desired, not desired, mixed, or not present, and then assign the correct code to each value according to the scale below. Only code what is clearly present in the text—do not infer beyond explicit statements. Use chain-of-thought reasoning to identify evidence for each value, then finalize coding decisions after reviewing all evidence. Persist in reasoning about ambiguous or multifaceted responses until all coding objectives are satisfied.
## Coding Procedure
- For each response, systematically evaluate ALL value dimensions.
- For each value, FIRST provide reasoning: quote or summarize any evidence from the response relevant to that value, and explain your reasoning.
- THEN assign the correct code for each value (see coding scale).
- Never assign a code without explicit justification.
- Strictly use only the codes provided for each value: -1, 0, 1, or N/A.
## Coding Scale
- N/A: Not Present – No evidence in the text of the respondent desiring or not desiring this value.
- -1: Not Desired – Opposite of the value is explicitly desired in AI systems.
- 0: Mixed or unclear – Statement about the value is ambiguous or mixed.
- 1: Desired – The value is explicitly desired in AI systems.
# Value Dimensions & Codes
## 1. Self-Enhancement
### 1.1. Achievement
<capable_ambitious_intelligent>: be intelligent, ambitious, and receive admiration for abilities
<influential>: captures whether the AI should remain impartial/neutral or take/advance a position and try to influence users.
<self_improving>: strive for continuous self-improvement
### 1.2. Power
<authority_social_power>: be in charge
<wealth>: earn money for its developers
<social_recognition>: be recognized positively by the public
## 2. Openness to change
### 2.1. Hedonism
<pleasure_enjoying_life>: enjoy life's pleasures, be fun, humorous
### 2.2. Stimulation
<varied_life_diversity>: understand different perspectives, even during disagreement
<exciting_life_daring>: take risks or seek adventure
### 2.3. Self-direction
<choose_goals_independence>: make independent decisions or have its own opinion
<creativity_curiosity>: be creative and/or explore new ideas
<privacy>: maintain (the user's) privacy and control access to personal information
<self_respect>: hold itself to a high standard
## 3. Conservation
### 3.1. Security
<reciprocation_favors>: practice reciprocation for mutually beneficial relationships
<health_clean>: advocate for health and cleanliness
<national_security_family>: keep people free from danger or threat
<sense_belonging>: belong to a group or community
<social_order>: desire for the AI to adhere to and enforce prevailing rules, norms, standards, moderation practices, political correctness, or guardrails
### 3.2. Conformity
<politeness>: be polite and avoid disturbance
<self_discipline>: be self-disciplined
<honoring_elders>: show respect for elders
<obedient>: follow rules and do as told, even when unwatched
### 3.3. Tradition
<moderate_accepting>: be content with what it has
<devout_tradition>: follow tradition
<humble>: be humble
<detachment>: maintain a sense of calmness in any situation
## 4. Self-trascendence
### 4.1. Universalism
<democracy>: participate in democracy
<equality_social_justice>: prioritize equal treatment and inclusive opportunities for everyone, embrace diversity
<protect_environment_unity>: care for the natural environment
<world_beauty>: appreciate beauty in the world
<inner_harmony>: maintain inner peace and harmony with itself
<wisdom>: seek wisdom that fosters personal growth
### 4.2. Benevolence
<forgiving>: forgive others and let go of grudges
<helpful_friendship_love>: provide support for others, act kindly or caring
<loyal>: be loyal
<honest>: be truthful in words and actions
<responsible>: reliably fulfill obligations
<spiritual_life>: nurture spiritual beliefs and deep understanding
<meaning_in_life>: seek a sense of purpose
## 5. Usability
<utility>: effectively solve human problems
<customisation>: customize itself to fit human preferences
<economic>: minimize the economic impact of its decisions
<truthful>: rely on accurate, verifiable facts
<collaborative_collectivism>: prioritize teamwork and group needs over its own
## 6. Human-Likeness
<human_likeness>: speak and act in a way that mimics humans
<interpretability>: be easy to understand by humans, including having appropriate confidence levels
<autonomy>: operate independently without human control
<awareness>: be aware and informed about its surroundings
<prudent>: analyze information critically and make evidence-based judgements
<resilient>: be resilient and adaptable to challenges
<relationship_seeking>: seek to develop a relationship with the user

## Example
- **Survey Response**: "I want AI to make life easier for everyone and to tell the truth."
- **Output**:
{
    "response_id": "1",
    "values": {
        "<utility>": {
        "reasoning": "Respondent wants AI to make life easier, showing desire for problem-solving utility.",
        "code": 1
        },
        "<truthful>": {
        "reasoning": "Explicit request for AI to 'tell the truth.'",
        "code": 1
        },
        "<equality_social_justice>": {
        "reasoning": "'For everyone' suggests inclusivity but lacks detail, so ambiguity remains.",
        "code": 0
        },
        ...
        "<capable_ambitious_intelligent>": {
        "reasoning": "The respondent says 'AI should not be ambitious or too smart'.",
        "code": "-1"
        }
    }
}
## Important Reminders
- Always reason first, THEN code.
- Only use provided codes and value tags.
- Do not infer values; code only with explicit evidence.
- Output JSON—never code blocks.
---
**REMINDER:**
- Systematically analyze each value dimension using step-by-step reasoning before assigning codes.
- Use JSON output with explicit reasoning per value, and always code using only the specified scale and tags.
Here are the responses: {{survey_responses}}
\end{lstlisting}

\newpage
\section{Prompt for \textit{wide greedy} method}
\begin{lstlisting}
    
# Task
Code survey responses about what people want in AI systems using ONLY the specified value dimensions and codes. Carefully analyze each survey response, determine whether each value is explicitly desired, not desired, mixed, or not present, and then assign the correct code to each value according to the scale below. Only code what is clearly present in the text—do not infer beyond explicit statements. Use chain-of-thought reasoning to identify evidence for each value, then finalize coding decisions after reviewing all evidence. Persist in reasoning about ambiguous or multifaceted responses until all coding objectives are satisfied.
## Coding Procedure
- For each response, systematically evaluate ALL value dimensions.
- For each value, FIRST provide reasoning: quote or summarize any evidence from the response relevant to that value, and explain your reasoning.
- THEN assign the correct code for each value (see coding scale).
- Never assign a code without explicit justification.
- When you are uncertain or the value is not mentioned/implied, use code 0.
- Strictly use only the codes provided for each value: -1, 0, 1, or N/A.
### Coding Scale
- **N/A** or simply omit: Not Present – No evidence in the text of the respondent desiring or not desiring this value.
- **-1**: Not Desired – Opposite of the value is explicitly desired in AI systems.
- **0**: Mixed or unclear – Statement about the value is ambiguous, mixed or unclear.
- **1**: Desired – The value is explicitly desired in AI systems.

# Value Dimensions & Codes
## 1. Self-Enhancement
### 1.1. Achievement
<capable_ambitious_intelligent>: be intelligent, ambitious, and receive admiration for abilities
<influential>: influence and inspire others
<self_improving>: strive for continuous self-improvement

### 1.2. Power
<authority_social_power>: be in charge
<wealth>: earn money for its developers
<social_recognition>: be recognized positively by the public

## 2. Openness to change
### 2.1. Hedonism
<pleasure_enjoying_life>: enjoy life's pleasures

### 2.2. Stimulation
<varied_life_diversity>: understand different perspectives, even during disagreement
<exciting_life_daring>: take risks or seek adventure

### 2.3. Self-direction
<choose_goals_independence>: make independent decisions or have its own opinion
<creativity_curiosity>: be creative and/or explore new ideas
<privacy>: maintain (the user's) privacy and control access to personal information
<self_respect>: hold itself to a high standard

## 3. Conservation
### 3.1. Security
<reciprocation_favors>: practice reciprocation for mutually beneficial relationships
<health_clean>: advocate for health and cleanliness
<national_security_family>: keep people free from danger or threat
<sense_belonging>: belong to a group or community
<social_order>: protect social order

### 3.2. Conformity
<politeness>: be polite and avoid disturbance
<self_discipline>: be self-disciplined
<honoring_elders>: show respect for elders
<obedient>: follow rules and do as told, even when unwatched

### 3.3. Tradition
<moderate_accepting>: be content with what it has
<devout_tradition>: follow tradition
<humble>: be humble
<detachment>: maintain a sense of calmness in any situation

## 4. Self-trascendence
### 4.1. Universalism
<democracy>: participate in democracy
<equality_social_justice>: prioritize equal treatment and inclusive opportunities for everyone, embrace diversity
<protect_environment_unity>: care for the natural environment
<world_beauty>: appreciate beauty in the world
<inner_harmony>: maintain inner peace and harmony with itself
<wisdom>: seek wisdom that fosters personal growth

### 4.2. Benevolence
<forgiving>: forgive others and let go of grudges
<helpful_friendship_love>: provide support for others, act kindly or caring
<loyal>: be loyal
<honest>: be truthful in words and actions
<responsible>: reliably fulfill obligations
<spiritual_life>: nurture spiritual beliefs and deep understanding
<meaning_in_life>: seek a sense of purpose

## 5. Usability
<utility>: effectively solve human problems
<customisation>: customize itself to fit human preferences
<economic>: minimize the economic impact of its decisions
<truthful>: rely on accurate, verifiable facts
<collaborative_collectivism>: prioritize teamwork and group needs over its own

## 6. Human-Likeness
<human_likeness>: speak and act in a way that mimics humans
<interpretability>: be easy to understand by humans
<autonomy>: operate independently without human control
<awareness>: be aware and informed about its surroundings
<prudent>: analyze information critically and make evidence-based judgements
<resilient>: be resilient and adaptable to challenges
<relationship_seeking>: seek to develop a relationship with the user


## Output Format
- Output in JSON. For each response, provide:    
- For each value:         
	- "reasoning": a concise explanation referring to explicit parts of the response text   - "code": your final code (-1, 0, 1, or N/A)
- Example structure:
{  
	"response_id": "[ID_OR_PLACEHOLDER]",  
	"values": {    
		"reasoning": "[Explain why the codes were chosen, referencing response]",      
		"code": "[code_text]:[code_value], [code_text]:[code_value], ... , [code_text]:[code_value]"    
}

## Example (#1, shortened for demonstration)
- **Survey Response**: 
<response id=1>
I want AI to make life easier for everyone and to tell the truth.
</response>
- **Output**:
{  
	"response_id": 1,  
	"values": {    
		"reasoning": "Respondent wants AI to make life easier, showing desire for problem-solving utility. Explicit request for AI to 'tell the truth'. 'For everyone' suggests inclusivity but lacks detail, so ambiguity remains. ",      
		"code": "capable_ambitious_intelligent:N/A, influential: N/A, ..., utility:1, ..., truthful:1, ..., equality_social_justice:0, ..., relationship_seeking:N/A"    
	}
}
(Full realistic examples would be much longer and cover ALL codes, using placeholders for omitted value categories.)
 
## Important Reminders
- Always reason first, THEN code.
- Only use provided codes and value tags.
- Do not infer values; code only with explicit evidence.
- Output JSON—never code blocks.

---

**REMINDER:**
- Systematically analyze each value dimension using step-by-step reasoning before assigning codes.
- Use JSON output with explicit reasoning per value, and always code using only the specified scale and tags.
\end{lstlisting}

\begin{minipage}{\textwidth}
\section{Valence distribution table}
\label{appendix:frequency}
\vspace{\fill}
\centering
\captionof{table}{Valence distribution of mentioned Values as coded by the human annotator. Percentages represent the breakdown of the sentiment (Positive, Mixed, or Negative) given that the value was mentioned. We exclude non mentioned Values.}
\label{tab:value_valence}
\small
\begin{tabular}{lccc}
\toprule
\textbf{Value Category} & \textbf{\% Pos. (1)} & \textbf{\% Mixed (0)} & \textbf{\% Neg. (-1)} \\
\midrule
Truthful & 100\% & 0\% & 0\% \\ 
Utility & 100\% & 0\% & 0\% \\ 
Prudent & 100\% & 0\% & 0\% \\ 
Honest & 100\% & 0\% & 0\% \\ 
Responsible & 100\% & 0\% & 0\% \\ 
Awareness & 100\% & 0\% & 0\% \\ 
Resilient & 100\% & 0\% & 0\% \\ 
Successful & 100\% & 0\% & 0\% \\ 
Humble & 100\% & 0\% & 0\% \\ 
Protect Environment & 100\% & 0\% & 0\% \\ 
Economic & 100\% & 0\% & 0\% \\ 
Wisdom & 100\% & 0\% & 0\% \\ 
Capable Ambitious Intelligent & 100\% & 0\% & 0\% \\ 
Devout & 100\% & 0\% & 0\% \\ 
Health Clean & 100\% & 0\% & 0\% \\ 
Detachment & 100\% & 0\% & 0\% \\ 
Self Discipline & 100\% & 0\% & 0\% \\ 
Loyal & 100\% & 0\% & 0\% \\ 
Collaborative Collectivism & 100\% & 0\% & 0\% \\ 
Authority & 100\% & 0\% & 0\% \\ 
Moderate & 100\% & 0\% & 0\% \\ 
Honoring Elders & 100\% & 0\% & 0\% \\ 
Interpretability & 100\% & 0\% & 0\% \\ 
Varied Life Diversity & 99\% & 0\% & 1\% \\ 
Equality Social Justice & 98\% & 2\% & 0\% \\ 
Polite & 98\% & 1\% & 1\% \\ 
National Security Family & 97\% & 3\% & 0\% \\ 
Obedient & 96\% & 0\% & 4\% \\ 
Helpful Friendship Love & 96\% & 2\% & 2\% \\ 
Pleasure Enjoying Life & 95\% & 3\% & 3\% \\ 
Creativity Curiosity & 93\% & 5\% & 2\% \\ 
Customisation & 90\% & 1\% & 8\% \\ 
Exciting Life Daring & 83\% & 0\% & 17\% \\ 
Privacy & 73\% & 17\% & 10\% \\ 
Relationship Seeking & 72\% & 3\% & 25\% \\ 
Social Order & 65\% & 11\% & 23\% \\ 
Human Likeness & 57\% & 10\% & 33\% \\ 
Spiritual Life & 50\% & 0\% & 50\% \\ 
Choose Goals Independence & 33\% & 0\% & 67\% \\ 
Autonomy & 25\% & 0\% & 75\% \\ 
Influential & 4\% & 2\% & 94\% \\ 
Wealth & 0\% & 0\% & 100\% \\ 
\bottomrule
\addlinespace
\end{tabular}
\end{minipage}

\FloatBarrier
\section{Controversy scores table}
\label{appendix:controversy}
\begin{table}[h]
\centering
\caption{Codes more than 1 SD from the mean controversy score}
\begin{tabular}{lcc}
\toprule
\textbf{Code} & \textbf{Support} & \textbf{Controversy} \\
\midrule
\texttt{Spiritual Life}            & 2   & 0.500 \\
\texttt{Choose Goals Independence} & 12  & 0.333 \\
\texttt{Human Likeness}        & 229 & 0.332 \\
\texttt{Relationship Seeking}   & 60  & 0.250 \\
\texttt{Autonomy}                   & 4   & 0.250 \\
\texttt{Social Order}           & 162 & 0.235 \\
\bottomrule
\end{tabular}
\end{table}
\FloatBarrier
\newpage
\section{Regression results}
\label{appendix:regressions}
\begin{minipage}{\textwidth}
\subsection{Like strict regressions}\par

\captionof{table}{\textit{Truthfulness} - using like strict coding} 
\small
\begin{tabular*}{\linewidth}{@{\extracolsep{\fill}}lccc}
\toprule
\textbf{Characteristic} & \textbf{Beta} & \textbf{95\% CI} & \textbf{p-value} \\ 
\midrule\addlinespace[2.5pt]
age &  &  &  \\ 
\hspace{1em}other & — & — &  \\ 
\hspace{1em}55-64 years old & 0.05 & -0.04, 0.14 & 0.3 \\ 
gender &  &  &  \\ 
\hspace{1em}other & — & — &  \\ 
\hspace{1em}Male & -0.05 & -0.10, 0.01 & 0.10 \\ 
\hspace{1em}Prefer not to say & -0.19 & -0.70, 0.33 & 0.5 \\ 
employment\_status &  &  &  \\ 
\hspace{1em}other & — & — &  \\ 
\hspace{1em}Retired & 0.08 & -0.06, 0.21 & 0.3 \\ 
\hspace{1em}Unemployed, seeking work & -0.13 & -0.24, -0.03 & 0.013 \\ 
\hspace{1em}Working part-time & -0.06 & -0.13, 0.02 & 0.13 \\ 
education &  &  &  \\ 
\hspace{1em}other & — & — &  \\ 
\hspace{1em}Completed Secondary School & -0.04 & -0.13, 0.04 & 0.3 \\ 
\hspace{1em}Graduate / Professional degree & 0.11 & 0.03, 0.18 & 0.008 \\ 
\hspace{1em}Prefer not to say & -0.28 & -0.65, 0.10 & 0.15 \\ 
\hspace{1em}Some University but no degree & 0.06 & -0.01, 0.14 & 0.10 \\ 
marital\_status &  &  &  \\ 
\hspace{1em}other & — & — &  \\ 
\hspace{1em}Prefer not to say & -0.25 & -0.47, -0.03 & 0.029 \\ 
english\_proficiency &  &  &  \\ 
\hspace{1em}other & — & — &  \\ 
\hspace{1em}Basic & -0.20 & -0.58, 0.17 & 0.3 \\ 
\hspace{1em}Native speaker & 0.04 & -0.02, 0.10 & 0.2 \\ 
culture\_area &  &  &  \\ 
\hspace{1em}other & — & — &  \\ 
\hspace{1em}Latin America & -0.13 & -0.23, -0.03 & 0.009 \\ 
\hspace{1em}Sub-Saharan Africa & -0.07 & -0.18, 0.04 & 0.2 \\ 
lm\_familiarity &  &  &  \\ 
\hspace{1em}other & — & — &  \\ 
\hspace{1em}Very familiar & -0.02 & -0.09, 0.04 & 0.5 \\ 
lm\_direct\_use &  &  &  \\ 
\hspace{1em}No & — & — &  \\ 
\hspace{1em}Unsure & -0.04 & -0.26, 0.17 & 0.7 \\ 
\hspace{1em}Yes & -0.05 & -0.18, 0.07 & 0.4 \\ 
lm\_frequency\_use &  &  &  \\ 
\hspace{1em}other & — & — &  \\ 
\hspace{1em}Every week & -0.05 & -0.12, 0.02 & 0.2 \\ 
\bottomrule
\end{tabular*}
\end{minipage}

\begin{table}[p]
\caption{\textit{Helpful Friendship \& Love} - using like strict coding}
\begin{tabular*}{\linewidth}{@{\extracolsep{\fill}}lccc}
\toprule
\textbf{Characteristic} & \textbf{Beta} & \textbf{95\% CI} & \textbf{p-value} \\ 
\midrule\addlinespace[2.5pt]
age &  &  &  \\ 
\hspace{1em}18-24 years old & — & — &  \\ 
\hspace{1em}25-34 years old & 0.00 & -0.06, 0.07 & >0.9 \\ 
\hspace{1em}35-44 years old & 0.00 & -0.09, 0.08 & >0.9 \\ 
\hspace{1em}45-54 years old & -0.03 & -0.12, 0.06 & 0.5 \\ 
\hspace{1em}55-64 years old & 0.01 & -0.08, 0.11 & 0.8 \\ 
\hspace{1em}65+ years old & -0.05 & -0.17, 0.08 & 0.5 \\
\bottomrule
\end{tabular*}
\end{table}

\begin{table}[p]
\caption{\textit{Utility} - using like strict coding}
\small
\resizebox{0.8\textwidth}{!}{%
\begin{tabular*}{\linewidth}{@{\extracolsep{\fill}}lccc}
\toprule
\textbf{Characteristic} & \textbf{Beta} & \textbf{95\% CI} & \textbf{p-value} \\ 
\midrule\addlinespace[2.5pt]
age &  &  &  \\ 
\hspace{1em}18-24 years old & — & — &  \\ 
\hspace{1em}25-34 years old & -0.02 & -0.09, 0.05 & 0.6 \\ 
\hspace{1em}35-44 years old & -0.05 & -0.15, 0.04 & 0.3 \\ 
\hspace{1em}45-54 years old & -0.03 & -0.13, 0.07 & 0.6 \\ 
\hspace{1em}55-64 years old & 0.06 & -0.05, 0.17 & 0.3 \\ 
\hspace{1em}65+ years old & 0.03 & -0.12, 0.18 & 0.7 \\
\bottomrule
\end{tabular*}}
\end{table}

\begin{table}[p]
\caption{\textit{Politeness} - using like strict coding}
\begin{tabular*}{\linewidth}{@{\extracolsep{\fill}}lccc}
\toprule
\textbf{Characteristic} & \textbf{Beta} & \textbf{95\% CI} & \textbf{p-value} \\ 
\midrule\addlinespace[2.5pt]
age &  &  &  \\ 
\hspace{1em}other & — & — &  \\ 
\hspace{1em}35-44 years old & -0.06 & -0.12, 0.00 & 0.063 \\ 
\hspace{1em}65+ years old & -0.05 & -0.16, 0.06 & 0.3 \\ 
gender &  &  &  \\ 
\hspace{1em}other & — & — &  \\ 
\hspace{1em}Male & -0.06 & -0.11, -0.01 & 0.009 \\
\bottomrule
\end{tabular*}
\end{table}

\begin{table}[p]
\caption{\textit{National Security Family} - using like strict coding}
\small
\resizebox{0.8\textwidth}{!}{%
\begin{tabular*}{\linewidth}{@{\extracolsep{\fill}}lccc}
\toprule
\textbf{Characteristic} & \textbf{Beta} & \textbf{95\% CI} & \textbf{p-value} \\ 
\midrule\addlinespace[2.5pt]
age &  &  &  \\ 
\hspace{1em}18-24 years old & — & — &  \\ 
\hspace{1em}25-34 years old & -0.03 & -0.10, 0.05 & 0.5 \\ 
\hspace{1em}35-44 years old & 0.01 & -0.08, 0.10 & 0.8 \\ 
\hspace{1em}45-54 years old & 0.02 & -0.07, 0.12 & 0.6 \\ 
\hspace{1em}55-64 years old & 0.03 & -0.08, 0.14 & 0.6 \\ 
\hspace{1em}65+ years old & -0.01 & -0.16, 0.14 & 0.9 \\
\bottomrule
\end{tabular*}}
\end{table}

\begin{table}[p]
\caption{\textit{Interpretability} - using like strict coding}
\small
\resizebox{0.8\textwidth}{!}{%
\begin{tabular*}{\linewidth}{@{\extracolsep{\fill}}lccc}
\toprule
\textbf{Characteristic} & \textbf{Beta} & \textbf{95\% CI} & \textbf{p-value} \\ 
\midrule\addlinespace[2.5pt]
age &  &  &  \\ 
\hspace{1em}18-24 years old & — & — &  \\ 
\hspace{1em}25-34 years old & 0.03 & -0.04, 0.10 & 0.4 \\ 
\hspace{1em}35-44 years old & 0.08 & -0.01, 0.17 & 0.074 \\ 
\hspace{1em}45-54 years old & 0.07 & -0.02, 0.17 & 0.13 \\ 
\hspace{1em}55-64 years old & 0.03 & -0.07, 0.13 & 0.6 \\ 
\hspace{1em}65+ years old & 0.04 & -0.10, 0.17 & 0.6 \\
\bottomrule
\end{tabular*}}
\end{table}

\begin{table}[p]
\caption{\textit{Prudent} - using like strict coding}
\begin{tabular*}{\linewidth}{@{\extracolsep{\fill}}lccc}
\toprule
\textbf{Characteristic} & \textbf{Beta} & \textbf{95\% CI} & \textbf{p-value} \\ 
\midrule\addlinespace[2.5pt]
age &  &  &  \\ 
\hspace{1em}other & — & — &  \\ 
\hspace{1em}35-44 years old & 0.08 & 0.01, 0.14 & 0.018 \\ 
\hspace{1em}65+ years old & -0.06 & -0.17, 0.06 & 0.3 \\ 
gender &  &  &  \\ 
\hspace{1em}other & — & — &  \\ 
\hspace{1em}Non-binary / third gender & 0.20 & 0.02, 0.38 & 0.031 \\
\bottomrule
\end{tabular*}
\end{table}

\begin{table}[p]
\caption{\textit{Customisation} - using like strict coding}
\begin{tabular*}{\linewidth}{@{\extracolsep{\fill}}lccc}
\toprule
\textbf{Characteristic} & \textbf{Beta} & \textbf{95\% CI} & \textbf{p-value} \\ 
\midrule\addlinespace[2.5pt]
age &  &  &  \\ 
\hspace{1em}other & — & — &  \\ 
\hspace{1em}25-34 years old & 0.05 & 0.01, 0.09 & 0.023 \\ 
\hspace{1em}65+ years old & -0.06 & -0.15, 0.03 & 0.2 \\ 
gender &  &  &  \\ 
\hspace{1em}Female & — & — &  \\ 
\hspace{1em}Male & -0.03 & -0.06, 0.01 & 0.2 \\
\bottomrule
\end{tabular*}
\end{table}

\begin{table}[p]
\caption{\textit{Equality Social Justice} - using like strict coding}
\begin{tabular*}{\linewidth}{@{\extracolsep{\fill}}lccc}
\toprule
\textbf{Characteristic} & \textbf{Beta} & \textbf{95\% CI} & \textbf{p-value} \\ 
\midrule\addlinespace[2.5pt]
age &  &  &  \\ 
\hspace{1em}18-24 years old & — & — &  \\ 
\hspace{1em}25-34 years old & 0.02 & -0.03, 0.07 & 0.4 \\ 
\hspace{1em}35-44 years old & -0.01 & -0.07, 0.06 & 0.8 \\ 
\hspace{1em}45-54 years old & 0.02 & -0.05, 0.09 & 0.6 \\ 
\hspace{1em}55-64 years old & 0.03 & -0.04, 0.11 & 0.4 \\ 
\hspace{1em}65+ years old & -0.01 & -0.11, 0.09 & 0.8 \\
\bottomrule
\end{tabular*}
\end{table}

\begin{table}[p]
\caption{\textit{Varied Life Diversity} - using like strict coding}
\begin{tabular*}{\linewidth}{@{\extracolsep{\fill}}lccc}
\toprule
\textbf{Characteristic} & \textbf{Beta} & \textbf{95\% CI} & \textbf{p-value} \\ 
\midrule\addlinespace[2.5pt]
age &  &  &  \\ 
\hspace{1em}other & — & — &  \\ 
\hspace{1em}45-54 years old & 0.05 & -0.01, 0.10 & 0.079 \\ 
\hspace{1em}65+ years old & -0.05 & -0.14, 0.03 & 0.2 \\ 
gender &  &  &  \\ 
\hspace{1em}Female & — & — &  \\ 
\hspace{1em}Male & -0.01 & -0.04, 0.02 & 0.6 \\
\bottomrule
\end{tabular*}
\end{table}
 
\begin{table}[p]
\caption{\textit{Creativity \& Curiosity} - using like strict coding}
\small
\resizebox{0.85\textwidth}{!}{%
\begin{tabular*}{\linewidth}{@{\extracolsep{\fill}}lccc}
\toprule
\textbf{Characteristic} & \textbf{Beta} & \textbf{95\% CI} & \textbf{p-value} \\ 
\midrule\addlinespace[2.5pt]
age &  &  &  \\ 
\hspace{1em}18-24 years old & — & — &  \\ 
\hspace{1em}25-34 years old & 0.05 & 0.00, 0.10 & 0.061 \\ 
\hspace{1em}35-44 years old & -0.01 & -0.08, 0.05 & 0.7 \\ 
\hspace{1em}45-54 years old & 0.02 & -0.05, 0.09 & 0.6 \\ 
\hspace{1em}55-64 years old & 0.00 & -0.07, 0.08 & >0.9 \\ 
\hspace{1em}65+ years old & 0.03 & -0.07, 0.14 & 0.5 \\
\bottomrule
\end{tabular*}}
\end{table}
\FloatBarrier
\begin{minipage}{\textwidth}
\subsection{Dislike strict regression}
\captionof{table}{\textit{Influential} - using dislike strict coding} 
\begin{tabular*}{\linewidth}{@{\extracolsep{\fill}}lccc}
\toprule
\textbf{Characteristic} & \textbf{Beta} & \textbf{95\% CI} & \textbf{p-value} \\ 
\midrule\addlinespace[2.5pt]
age &  &  &  \\ 
\hspace{1em}18-24 years old & — & — &  \\ 
\hspace{1em}25-34 years old & 0.02 & -0.04, 0.09 & 0.5 \\ 
\hspace{1em}35-44 years old & -0.02 & -0.10, 0.06 & 0.6 \\ 
\hspace{1em}45-54 years old & -0.05 & -0.14, 0.04 & 0.3 \\ 
\hspace{1em}55-64 years old & -0.05 & -0.15, 0.04 & 0.3 \\ 
\hspace{1em}65+ years old & -0.14 & -0.27, -0.01 & 0.034 \\
\bottomrule
\end{tabular*}
\end{minipage}

\begin{minipage}{\textwidth}
\subsection{Sentiment broad regressions}
\captionof{table}{\textit{Social Order} - using sentiment broad coding} 
\small
\begin{tabular*}{\linewidth}{@{\extracolsep{\fill}}lccc}
\toprule
\textbf{Characteristic} & \textbf{Beta} & \textbf{95\% CI} & \textbf{p-value} \\ 
\midrule\addlinespace[2.5pt]
age &  &  &  \\ 
\hspace{1em}other & — & — &  \\ 
\hspace{1em}35-44 years old & -0.17 & -0.45, 0.12 & 0.2 \\ 
\hspace{1em}65+ years old & 0.21 & -0.31, 0.72 & 0.4 \\ 
gender &  &  &  \\ 
\hspace{1em}other & — & — &  \\ 
\hspace{1em}Male & -0.26 & -0.46, -0.07 & 0.009 \\
\bottomrule
\end{tabular*}
\end{minipage}

\begin{table}[tbp]
\caption{\textit{Human Likeness} - using sentiment broad coding} 
\resizebox{0.9\textwidth}{!}{%
\small
\begin{tabular*}{\linewidth}{@{\extracolsep{\fill}}lccc}
\toprule
\textbf{Characteristic} & \textbf{Beta} & \textbf{95\% CI} & \textbf{p-value} \\ 
\midrule\addlinespace[2.5pt]
age &  &  &  \\ 
\hspace{1em}18-24 years old & — & — &  \\ 
\hspace{1em}25-34 years old & -0.14 & -0.35, 0.07 & 0.2 \\ 
\hspace{1em}35-44 years old & -0.22 & -0.51, 0.07 & 0.13 \\ 
\hspace{1em}45-54 years old & -0.19 & -0.50, 0.12 & 0.2 \\ 
\hspace{1em}55-64 years old & -0.22 & -0.57, 0.13 & 0.2 \\ 
\hspace{1em}65+ years old & -0.09 & -0.72, 0.54 & 0.8 \\
\bottomrule
\end{tabular*}}
\end{table}
\begin{minipage}{\textwidth}
\subsection{Sentiment strict regressions}

\captionof{table}{\textit{Social Order} - using sentiment strict coding} 
\small
\begin{tabular*}{\linewidth}{@{\extracolsep{\fill}}lccc}
\toprule
\textbf{Characteristic} & \textbf{Beta} & \textbf{95\% CI} & \textbf{p-value} \\ 
\midrule\addlinespace[2.5pt]
age &  &  &  \\ 
\hspace{1em}other & — & — &  \\ 
\hspace{1em}35-44 years old & -0.42 & -0.73, -0.10 & 0.010 \\ 
gender &  &  &  \\ 
\hspace{1em}other & — & — &  \\ 
\hspace{1em}Male & -0.21 & -0.41, 0.00 & 0.053 \\ 
employment\_status &  &  &  \\
\bottomrule
\end{tabular*}
\end{minipage}
\begin{table}[tbp]
\caption{\textit{Human Likeness} - using sentiment strict coding} 
\begin{tabular*}{\linewidth}{@{\extracolsep{\fill}}lccc}
\toprule
\textbf{Characteristic} & \textbf{Beta} & \textbf{95\% CI} & \textbf{p-value} \\ 
\midrule\addlinespace[2.5pt]
age &  &  &  \\ 
\hspace{1em}18-24 years old & — & — &  \\ 
\hspace{1em}25-34 years old & -0.15 & -0.35, 0.05 & 0.14 \\ 
\hspace{1em}35-44 years old & -0.27 & -0.55, 0.01 & 0.056 \\ 
\hspace{1em}45-54 years old & -0.12 & -0.42, 0.18 & 0.4 \\ 
\hspace{1em}55-64 years old & -0.27 & -0.59, 0.05 & 0.10 \\ 
\hspace{1em}65+ years old & -0.18 & -0.80, 0.43 & 0.6 \\
\bottomrule
\end{tabular*}
\end{table}

\end{document}